\newcommand{\CLASSINPUTinnersidemargin}{0.75in} 
\newcommand{\CLASSINPUToutersidemargin}{0.75in} 
\newcommand{\CLASSINPUTtoptextmargin}{0.75in} 
\newcommand{\CLASSINPUTbottomtextmargin}{0.75in}
\documentclass[letterpaper, 10 pt, conference]{IEEEtran}

\IEEEoverridecommandlockouts

\usepackage[normalem]{ulem}
\usepackage[table,usenames,dvipsnames]{xcolor}
\usepackage{extarrows}
\let\labelindent\relax
\usepackage[inline]{enumitem}
\usepackage{cite}
\usepackage{lipsum}
\usepackage[dvipsnames]{xcolor}

\usepackage{amsmath,amssymb,amsfonts, amsthm, dsfont} 
\usepackage{mathtools, bm}
\usepackage[linesnumbered, lined, boxed, ruled]{algorithm2e}

\usepackage{graphicx}
\usepackage{tabularx}
\usepackage[font={small}]{caption}   
\usepackage{subcaption}
\usepackage[breaklinks=true, colorlinks, bookmarks=true, citecolor=Black, urlcolor=Violet,linkcolor=Black]{hyperref}

\def\argmax{\mathop{\arg\max}\limits}	%

\title{\vspace{6mm}\LARGE \bf
Stochastic 2-D Motion Planning with a POMDP Framework
}

\author{
Ke Sun and Vijay Kumar*
\thanks{ We gratefully acknowledge the support of ARL grants W911NF-08-2-0004, W911NF-17-2-0181 and DARPA grants HR001151626, HR0011516850.}
\thanks{*The authors are with GRASP Lab, University of Pennsylvania, Philadelphia, PA 19104, USA, {\tt\small\{sunke, kumar\}@seas.upenn.edu}.}
}

\begin{document}

\maketitle
\thispagestyle{empty}
\pagestyle{empty}

\begin{abstract}
  Motion planning is challenging when it comes to the case of imperfect state information. Decision should be made based on belief state which evolves according to the noise from the system dynamics and sensor measurement. In this paper, we propose the QV-Tree Search algorithm which combines the state-of-art offline and online approximation methods for POMDP. Instead of full node expansions in the tree search, only probable future observations are considered through forward sampling. This modification helps reduce online computation time and allows for GPU acceleration. We show using representative examples that the proposed QV-Tree Search is able to actively localize the robot in order to reach the goal location with high probability. The results of the proposed method is also compared with the A* and MDP algorithms, neither of which handles state uncertainty directly. The comparison shows that QV-Tree Search is able to drive the robot to the goal with higher success rate and fewer steps.
\end{abstract}

\section{Introduction}
\label{sec: introduction}
Navigation through complex environment is one of the most important modules in achieving full robot autonomy. The problem is embeded in many applications of robots such as coastal navigation~\cite{roy2000coastal}, trajectory planning for mobile vehicles~\cite{costante2018exploiting, achtelik2014motion}, and robot grasping~\cite{platt2017efficient}. In order to navigate robustly in the real world environment, a planning algorithm should be able to handle noise from different sources including robot dynamics, sensor measurements, and external environment. In this paper, we will focus on developing a motion planning algorithm handling uncertainty from the first two sources assuming a static environment known as \textit{a priori}.

Based on whether an algorithm assumes dynamics or measurement noise, most of the planning algorithms can be categorized into the following three groups. With the assumption of deterministic motion and perfect state information, the motion planning problem is well solved in both discrete and continuous space with algorithms like A*~\cite{hart1972correction}, RRT*~\cite{karaman2011sampling}, and PRM~\cite{kavraki1996probabilistic}.
If the state and action spaces are discrete, motion planning with stochastic motion execution can be readily modeled as a Markov Decision Process (MDP)~\cite[Ch.5]{bertsekas1995dynamic}. For continuous state space, Melchior and Simmons~\cite{melchior2007particle} and Alterovitz, \textit{et al}.~\cite{alterovitz2007stochastic} extend the work of RRT and PRB. Khatib~\cite{khatib1986real} also proposes the application of artificial potential field so that a robot can maintain a safe path through minimizing the associated energy.
Although the state estimation technics advances significantly in recent years~\cite{zhang2014loam, sun2018robust}, the assumption of perfect state information is still easily violated in practice affecting the robustness of the robot autonomy. The fragility of the state estimation is usually due to its underlying observability~\cite{martinelli2012vision}. Because of the nonlinear nature of most state estimation algorithms, solving such problems requires carefully designed trajectories. The idea movtivates works on planning in belief space~\cite{prentice2009belief, bry2011rapidly, van2011lqg, vitus2011closed, indelman2015planning}. These works share the assumption that the system is linear with noise of Gaussian distribution. The benefit is that the posterior belief of the state can be computed in closed form and is independent with the actual future measurement. However, for nonlinear measurement models, the estimated future belief can be arbitrarily inaccurate. Platt \textit{et al}.~\cite{platt2017efficient} propose to model the non-Gaussian belief space with a state hypothesis and additional state samples. Since the actual future measurement will affect the belief evolution now, the measurement with maximum likelihood is assumed~\cite{platt2010belief}.

Planning for systems with stochastic motion and measurement can be naturally modelled by Partrially Observable Markov Decision Process (POMDP), which is a well studied field in the AI community~\cite[Ch.4]{bertsekas1995dynamic}. Especially for discrete systems with finite state, action, and observation space, Smallwood and Sondik~\cite{smallwood1973optimal} show that the optimal value function is a piecewise linear convex function of the belief, and can be solved through value iteration. Although refined by Kaelbling \textit{et al}.~\cite{kaelbling1998planning} and Zhang \textit{et al}.~\cite{zhang2001speeding}, the POMDP framework is still not widely used in the planning mainly due to the exponentially growing computational complexity. Efforts are made to approximate the optimal value function through either point-based methods~\cite{pineau2003point, hauskrecht2000value, kurniawati2008sarsop, smith2012point, shani2013survey} or finite state machine~\cite{kaelbling1998planning, pajarinen2011periodic}, which restricts belief or policy space respectively. Besides offline approximation methods, there are tree search~\cite{browne2012survey} based online methods~\cite{ross2007aems, washington1997bi, ross2008online} to further improve the approximation accuracy of the value function. Most of the online methods are different in what heuristics is applied when selecting the next leaf node to expand. Recent works from Silver~\cite{silver2010monte} and Cai~\cite{Cai-RSS-18} combine tree search with sampling-based methods to deal with large scale problems.

In this paper, we propose QV-Tree Search (QVTS), named after the two types of nodes in the tree structure, to solve the 2-D discrete motion planning problem in the POMDP framework, which considers both dynamics and measurement noise in the system model. We take advantage of both state-of-art offline and online POMDP approximation methods to increase the value approximation accuracy of the current belief. Experiments show that less than $1.5s$ is required in stepwise planning for state space of size $4\times10^3$ with the laptop-level CPU and GPU. To the best of our knowledge, this is the first time that the motion planning problem of practical size is solved in the POMDP framework within a reasonable online computation time. An implementation of the proposed work is open sourced at \url{https://github.com/KumarRobotics/path_planning_2d}.

The rest of the paper is organized as follows: Sec.~\ref{sec: problem formulation} will give a formal definition of the problem followed by necessary POMDP preliminaries in Sec.~\ref{sec: pomdp preliminary}. A detailed description of the QVTS is presented in Sec.~\ref{sec: qv tree search}. In Sec.~\ref{sec: experiments}, we first provide a simple example applying the proposed algorithm. Then we compare the performance with other widely used motion planning algorithms including A* and MDP. Sec.~\ref{sec: conclusion} concludes the paper.

\section{Problem Formulation}
\label{sec: problem formulation}
A motion planning problem can be described by the following discrete (discrete-time, discrete-state) model,
\begin{equation*}
  \begin{aligned}
    x_{k+1} &= f(x_k, a_k , w_k), \\
    z_k &= h(x_k, v_k),
  \end{aligned}
\end{equation*}
where $x_k\in\mathcal{X}$, $a_k\in\mathcal{A}$, and $z_k\in\mathcal{Z}$ for $k = 0, 1, 2,\dots$ are the states, actions and observations of a robot with $\mathcal{X}$, $\mathcal{A}$, and $\mathcal{Z}$ being finite sets. $w_k$ and $v_k$ model the transition and measurement uncertainty, i.e. $P(x_{k+1}|x_k, a_k) = P(w_k|x_k, a_k)$ and $P(z_k|x_k) = P(v_k|x_k)$. In the problem, we assume the robot cannot access the state information directly but through sensor measurement with initial distribution defined as $b_0 \coloneqq P(x_0)\in\mathcal{B}(\mathcal{X})$.
Given a known occupancy grid map $m:\mathcal{X} \mapsto \{0, 1\}$, one can naturally define a stage reward function $R:\mathcal{X}\times\mathcal{A}\mapsto\mathbb{R}$. The goal is to find an optimal admissible stationary policy $\mu^*:\mathcal{B}\mapsto\mathcal{A}$ which maximizes the reward obtained for an infinite horizon,
\begin{equation}
  \label{eq: value function}
  V_{\mu}(b_0) =
  \underset{\substack{x_0, w_k, v_k\\k=0,1,2\dots}}{\mathbb{E}}
  \left\{\sum_{k=0}^\infty \gamma^k R\left(x_k, \mu(b_k)\right)\right\},
\end{equation}
where $\gamma\in(0, 1)$ is a discount factor to ensure the summation is finite. Intuitively, the policy should be a function of an information vector, i.e. all information including actions and measurements in the history. Smallwood and Sondik~\cite{smallwood1973optimal} show that the belief state can serve as the sufficient statistics, i.e. the policy can also be a function of the belief state.

More concisely, the model data can be summarized with the tuple $(\mathcal{X}, \mathcal{A}, \mathcal{Z}, T, O, R, b_0, \gamma)$, where $T:\mathcal{X}\times\mathcal{A}\times\mathcal{X}\mapsto[0, 1]$ and $O:\mathcal{X}\times\mathcal{Z}\mapsto[0, 1]$ are the transition probability and measurement likelihood with the rest of elements defined as above. Note that the problem formulation fits stochastic motion planning problems of arbitrary dimension. However, this paper will focus on the 2-D motion planning problem.

\section{POMDP Preliminary}
\label{sec: pomdp preliminary}
Based on the problem formulation, the stochastic motion planning problem fits naturally in a POMDP framework. As shown in~\cite[Ch.4]{bertsekas1995dynamic}, the optimal value function $V^*$ of the optimal policy $\mu^*$ is the fixed point of,
\begin{equation}
  \label{eq: pomdp bellman equation}
  \begin{aligned}
    V^*(b) &= \max_{a\in\mathcal{A}} \left[ R(b, a) +
    \gamma \sum_{z\in\mathcal{Z}} P(z|b, a) V^*(\Phi(b, a, z)) \right] \\
    &=  R(b, \mu^*(b)) + \gamma \sum_{z\in\mathcal{Z}}
    P(z|b, \mu^*(b)) V^*(\Phi(b, \mu^*(b), z)),
  \end{aligned}
\end{equation}
where $R(b, a) \coloneqq \sum_{x\in\mathcal{X}} R(x, a) b(x)$ and $\Phi: \mathcal{B}\times\mathcal{A}\times\mathcal{Z} \mapsto \mathcal{B}$ represent the Bayes belief update,
\begin{equation}
  \label{eq: posterior belief update}
  \Phi(b, a, z)(x') = \frac{O(x', z)}{P(z|b, a)}
  \sum_{x\in\mathcal{X}} T(x, a, x')b(x).
\end{equation}
Especially, for problems with finite states, Smallwood and Sondik~\cite{smallwood1973optimal} show that the optimal value function is a piecewise linear convex function of the belief state, i.e.
\begin{equation}
  \label{eq: piecewise linear convex value function}
  V^*(b) = \max_{\alpha\in\Gamma^*} \alpha^\top b,
\end{equation}
where each $\alpha$-vector in the set $\Gamma^*$ is an $|\mathcal{X}|$ dimensional hyperplane defining $V^*$ over a bounded region over the belief state. Each $\alpha$-vector is also associated with a specific action. The special structure of the optimal value function allows Eq.~\eqref{eq: pomdp bellman equation} to be rewritten as,
\begin{equation}
  \begin{gathered}
    \label{eq: finite state pomdp bellman equation}
    V^*(b) = \max_{a\in\mathcal{A}} \bigg[ R(b, a) + \\
      \gamma \sum_{z\in\mathcal{Z}} \max_{\alpha\in\Gamma^*}
      \sum_{x\in\mathcal{X}}\sum_{x'\in\mathcal{X'}}
      O(x', z)T(x, a, x')\alpha(x')b(x) \bigg].
  \end{gathered}
\end{equation}
However, computing the exact $\Gamma^*$ can be computationally prohibitive. More precisely, if computed with value iteration, in the worst case, the number of piecewise linear regions at iteration $k+1$ is $|\Gamma_{k+1}| = |\mathcal{A}||\Gamma_k|^{|\mathcal{Z}|}$.

\section{QV-Tree Search}
\label{sec: qv tree search}
In this paper, we propose QV-Tree Search (QVTS) adapted from the general tree search method to find the optimal action online. QVTS also takes advantages of the existing offline approximations, Fast Informed Bound (FIB)~\cite{hauskrecht2000value} and Point-based Value Iteration~\cite{shani2013survey}. Note that the framework of QVTS allows easy swap of the offline approximation algorithms when appropriate. The two algorithms chosen in our work are know to be the most accurate algorithms in estimating the upper and lower bound of the reward-to-go for the POMDP problems. Also, comparing with using Monte Carlo simulation to estimate the reward-to-go~\cite{silver2010monte, Cai-RSS-18}, using the results of offline methods directly is superior in both computation speed and estimation accuracy. In the following subsections, we will briefly summarize the FIB and PBVI algorithms from~\cite{hauskrecht2000value} and~\cite{shani2013survey}, and then explain the QVTS in detail.

\subsection{Fast Informed Bound}
\label{subsec: fast informed bound}
Most of the approximation methods for POMDP exploit the fact that the optimal value function is piecewise linear over the belief state. Instead of having possibly infinite number of piecewise linear regions, such methods use a fixed set of $\alpha$-vectors $\Gamma$, where each element $\alpha \in \mathbb{R}^{|\mathcal{X}|}$ is associated with an action.
FIB~\cite{hauskrecht2000value} is special in that it assigns one single $\alpha$-vector for each action $a\in\mathcal{A}$, i.e. $|\Gamma_{FIB}| = |\mathcal{A}|$. 
\begin{equation}
  \begin{gathered}
    \label{eq: fib bellman equation}
    V_{FIB}(b) = \max_{a\in\mathcal{A}} \bigg[ R(b, a) + \\
      \gamma \sum_{z\in\mathcal{Z}} \sum_{x\in\mathcal{X}}
      \max_{\alpha\in\Gamma_{FIB}}\sum_{x'\in\mathcal{X'}}
      O(x', z)T(x, a, x')\alpha(x')b(x) \bigg].
  \end{gathered}
\end{equation}
Instead of using Eq.~\eqref{eq: fib bellman equation}, one can write out the ``Bellman equation'' for the $\alpha$-vectors directly,
\begin{equation}
  \label{eq: fib alpha vector bellman equation}
  \begin{gathered}
    \alpha^a_{FIB}(x) = R(x, a) + \\
    \gamma \sum_{z\in\mathcal{Z}} \max_{\alpha^{a'}_{FIB}\in\Gamma_{FIB}}
    \sum_{x'\in\mathcal{X}} O(x', z) T(x, a, x') \alpha^{a'}_{FIB}(x'),
  \end{gathered}
\end{equation}
FIB is advantagous in that Eq.~\eqref{eq: fib alpha vector bellman equation} is a monotonic and contractive mapping. Therefore, the $\alpha$-vectors converges by applying value iteration with Eq.~\eqref{eq: fib alpha vector bellman equation} with the complexity of each iteration $O(|\mathcal{A}|^2|\mathcal{X}|^2|\mathcal{Z}|)$.

\subsection{Point-based Value Iteration}
\label{subsec: point based value iteration}
Given an initial belief, $b_0$, an intuitive thought is that most of regions in the belief space will not be reached with an arbitrary sequence of actions and observations. Therefore, instead of treating every point in the belief space equally, more computation effort should spend on the points that are more probable to reach.
PBVI assumes that a belief set $\mathcal{B}_{PBVI}$ is given, where $b_0\in\mathcal{B}_{PBVI}$. Each point $b\in\mathcal{B}_{PBVI}$ is associated with an $\alpha$-vector, i.e. $|\Gamma_{PBVI}| = |\mathcal{B}_{PBVI}|$. The ``Bellman equation'' for the $\alpha$-vectors follows,
\begin{equation}
  \begin{gathered}
    \label{eq: pbvi alpha vector bellman equation}
    V_{PBVI}(b) = \max_{a\in\mathcal{A}} \bigg[ R(b, a) + \\
      \gamma \sum_{z\in\mathcal{Z}} \max_{\alpha\in\Gamma_{PBVI}}
      \sum_{x\in\mathcal{X}}\sum_{x'\in\mathcal{X'}}
      O(x', z)T(x, a, x')\alpha(x')b(x) \bigg], \\
      \forall b \in \mathcal{B}_{PBVI}.
  \end{gathered}
\end{equation}
Although there is no difference between Eq.~\eqref{eq: pbvi alpha vector bellman equation} and the true Bellman equation~\eqref{eq: finite state pomdp bellman equation}, Eq.~\eqref{eq: pbvi alpha vector bellman equation} only considers points in $\mathcal{B}_{PBVI}$ instead of the entire belief space $\mathcal{B}$.

Different point-based approximation methods~\cite{pineau2003point, smith2004heuristic, kurniawati2008sarsop, smith2012point, shani2013survey} differ in how the reachable set $\mathcal{B}_{PBVI}$ is generated. In the proposed method, we follow the belief set expansion algorithm introduced in~\cite{pineau2003point} because of its implementation simplicity. At each iteration in the expansion, a new belief point is generated using every point already in the set by forward sampling an action-observation trajectory.

The final set of $\alpha$-vectors can also be obtained through value iteration using Eq.~\eqref{eq: pbvi alpha vector bellman equation} with each iteration having complexity $O(|\mathcal{X}||\mathcal{A}||\mathcal{Z}||\mathcal{B}|^2)$. Unlike FIB, value iteration may not lead to a convergent $V_{PBVI}$ with point-based methods. However, the error $\|V_{PBVI}-V^*\|_\infty$ is bounded proved by Pineau \textit{et al.}~\cite{pineau2003point}. And the bound gets tighter as the density of the belief set increase.

\subsection{Online Tree Search}
\label{subsec: online tree search}
\begin{figure}[tb]
  \centering
  \includegraphics[scale=0.2]{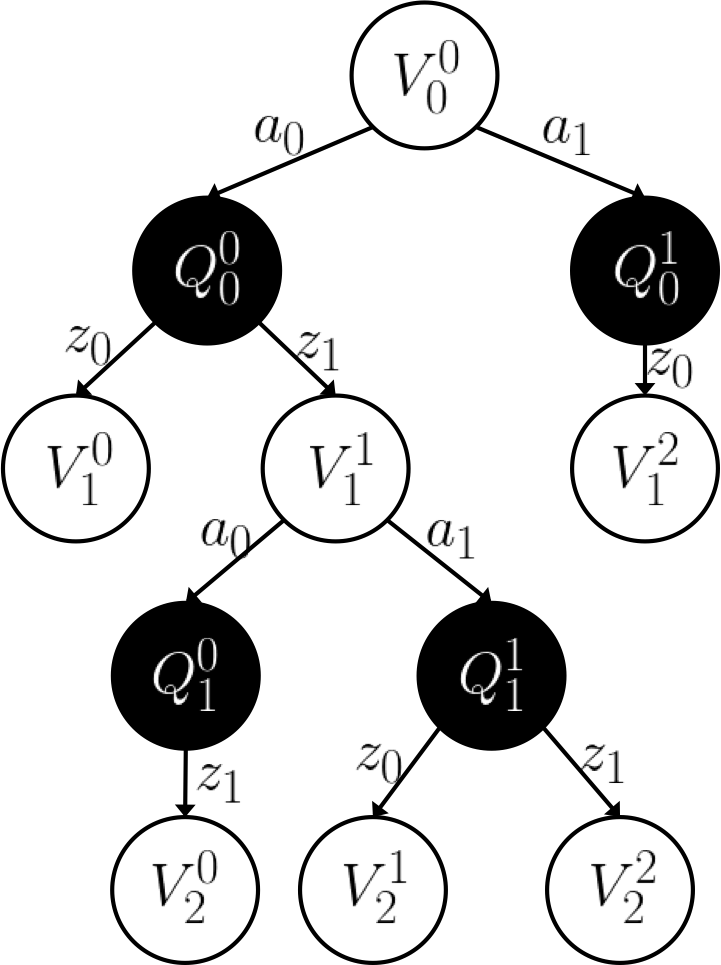}
  \caption{An example of QV-Tree with $|\mathcal{A}|=|\mathcal{Z}|=2$. Note that all actions are considered during a node expansion. However, only the probable observations are considered in order to save online computation time.}
  \label{fig: example qv search tree}
\end{figure}

The approximation methods introduced in Sec.~\ref{subsec: fast informed bound} and~\ref{subsec: point based value iteration} can be directly used, as Eq.~\eqref{eq: piecewise linear convex value function}, online to estimate the value and find the corresponding action. In order to further improve the value estimation accuracy, we propose QVTS combining the above two methods and tree search untilizing the online computation time for multi-step lookahead. Since tree search framework allows termination based on time constraint, the proposed algorithm is an anytime algorithm for POMDP.
\begin{algorithm}[t]
  \caption{General framework for QV-Tree Search}
  \label{alg: general framework for qv-tree search}
  \SetKwInOut{Input}{Input}
  \SetKwFunction{QVSearchTree}{QVSearchTree}
  \SetKwFunction{taskFinished}{taskFinished}
  \SetKwFunction{planningFinished}{planningFinished}
  \SetKwFunction{initialize}{initialize}
  \SetKwFunction{findVNodeToExpand}{findVNodeToExpand}
  \SetKwFunction{expand}{expand}
  \SetKwFunction{root}{root}
  \SetKwFunction{update}{update}
  \SetKwFunction{getOptimalAction}{getOptimalAction}

  \Input{$b_0$}

  $s = $ \QVSearchTree{$b_0$}\;
  \While{not \taskFinished{}}{
    \While{not \planningFinished{}}{
      $v$ = $s$.\findVNodeToExpand{}\;
      $v$.\expand{}\;
      $p$ = $v$.$parent$\;
      \tcp{$s.root.parent = NULL$}
      \While{$p \neq$ NULL}{
        $p$.\update{}\;
        $p$ = $p$.$parent$\;
      }
    }
    $a$ = $s$.\getOptimalAction{}\;
    Execute action $a$\;
    Receive an measurement $z$\;
    $s$.\update{$a$, $z$}\;
  }
\end{algorithm}
Unlike MDP problems, the QV-Tree structure includes two types of nodes, $Q$-nodes and $V$-nodes, interleaving each other at different layers of the tree. As the name suggests, the $Q$-nodes are functions of the belief-action pair as $q$-functions, while the $V$-nodes are functions of the belief only as value functions. Figure~\ref{fig: example qv search tree} shows an example of a QV-Tree.
One of the core idea of tree search is to expand the most ``promising'' leaf node in order to save computation. The same idea is achieved through associating four variables with each node, including an upper bound $U_x$, a lower bound $L_x$, a heuristic value $H_x$, and a $V$-node to expand $E_x$ within the subtree~\cite{satia1973markovian, washington1997bi, ross2007aems}. The values combined indicate how much the approximation error at the node would affect the value estimated at the root. The heuristics at the $V$-nodes and $Q$-nodes are defined as follows,
\begin{equation}
  \label{eq: heuristic value}
  \begin{gathered}
    H_V(b) =
    \begin{cases}
      V_{FIB}(b) - V_{PBVI}(b), & \text{if $V$ is a leaf node}. \\
      \max_{a\in\mathcal{A}} H(b, a) H_Q(b, a), & \text{otherwise}.
    \end{cases} \\
    H_Q(b, a) = \max_{z\in\mathcal{Z}} H(b, a, z) H_V(\Phi(b, a, z)).
  \end{gathered}
\end{equation}
In our implementation, we follow the heuristic propogation method proposed in~\cite{ross2007aems}, where
\begin{equation*}
  \begin{gathered}
    H(b, a) =
    \begin{cases}
      1, & \text{if } a = \argmax_{a'\in\mathcal{A}} U_Q(b, a'). \\
      0, & \text{otherwise}.
    \end{cases} \\
    H(b, a, z) = \gamma P(z|b, a).
  \end{gathered}
\end{equation*}

Algorithm~\ref{alg: general framework for qv-tree search} shows the general framework to construct and update a QV-Tree. Since each node contains the $V$-node to expand within its subtree, specially, the root node will point to the best $V$-node to expand of the entire QV-Tree. Therefore, \verb!findVNodeToExpand()! should trivially return the $V$-node pointed by the root. After expanding the selected $V$-node, the infomation of all the parent nodes up to the root node will be updated. When the planning terminates, either the planning time constraint is met or the value-to-go estimated at the root node is accurate enough, the robot selects the optimal action based on the current QV-Tree. Knowing the executed action, together with the newly received measurement, the QV-Tree is updated and set the root to be consistent with the current belief. The following introduces the details of the node expansion and tree update.

\begin{algorithm}
  \caption{$V$-node Expansion}
  \label{alg: vnode expansion}
  \SetKwFunction{qnodeConstruction}{QNode}
  \SetKwFunction{append}{append}
  \SetKwFunction{update}{update}

  \ForEach{$a\in\mathcal{A}$}{
    $q =$ \qnodeConstruction{$v.belief$, $a$}\;
    $v.children$.\append{$q$}\;
  }
  $v$.\update{}\; Also, Also,
\end{algorithm}
The expansion of a $V$-node can be logically separated into Algorithm~\ref{alg: vnode expansion}-\ref{alg: vnode update}. In general, expanding a $V$-node starts with creating $|\mathcal{A}|$ child $Q$-nodes. For each child $Q$-node, $\mathcal{Z}$ new leaf $V$-nodes should be created in order to cover all possible measurements at the next step. The creation of $|\mathcal{A}|$ $Q$-nodes is necessary in order to consider all possible policies. However, instead of explicitly expanding all measurements for the $Q$-nodes, measurement samples are generated through forward sampling given the current belief and the action taken as in Algorithm~\ref{alg: forward sampling}. \begin{algorithm}
  \caption{$Q$-node Construction}
  \label{alg: qnode construction}
  \SetKwInOut{Input}{Input}
  \SetKwInOut{Output}{Output}
  \SetKwFunction{forwardSampling}{\textbf{forwardSampling}}
  \SetKwFunction{beliefUpdate}{\textbf{beliefUpdate}}
  \SetKwFunction{vnodeConstruction}{VNode}
  \SetKwFunction{append}{append}
  \SetKwFunction{update}{update}

  \Input{$b$, $a$, $v$}
  \Output{$q$}

  \tcp{Functions with bold font run on GPU.}
  $q.belief = b$\;
  $q.action = a$\;
  $q.parent = v$\;
  \tcp{Generate $n$ sample measurements.}
  $\mathcal{M} =$ \forwardSampling{$b$, $a$, $n$}\;
  Get the unique measurements $\bar{\mathcal{M}}$ and the frequency $\bar{\mathcal{F}}$\;
  \ForEach{$m, f\in \bar{\mathcal{M}}, \bar{\mathcal{F}}$}{
    $b' =$ \beliefUpdate{$b$, $a$, $m$}\;
    $v =$ \vnodeConstruction{$b'$, $m$, $f$, $q$}\;
    $q.children$.\append{$v$}\;
  }
  $q$.\update{}\;
\end{algorithm}
\begin{algorithm}
  \caption{Forward Sampling}
  \label{alg: forward sampling}

  \SetKwInOut{Input}{Input}
  \SetKwInOut{Output}{Output}

  \Input{$b$, $a$, $n$}
  \Output{$\mathcal{M}$}

  $\mathcal{M} = \emptyset$\;
  \ForEach{thread from $1$ to $n$}{
    Sample $x \sim b$\;
    Sample $x' \sim P(x'|x, a)$\;
    Sample $z \sim P(z|x')$\;
    $\mathcal{M} = \mathcal{M}\cup \{z\}$\;
  }
\end{algorithm}
The benefit of forward sampling is two-fold. In the case that $|\mathcal{Z}|$ is large, most of the measurements are unlikely to happen, i.e. $P(z|b, a)\approx 0$. As indicated in Eq.~\eqref{eq: pomdp bellman equation}, the $V$-nodes with such measurements will have little effect on the approximation accuracy of the root node. Also, it is expensive to compute the marginal probability of a measurement $P(z|b, a)$ and the posterior belief as in Eq.~\eqref{eq: posterior belief update}. Replacing the full expansion of a $Q$-node with forward sampling is more efficient in that forward sampling will implicitly ignore the measurements with small $P(z|b, a)$. The second benefit is that forward sampling is a simple linear procedure as listed in Algorithm~\ref{alg: forward sampling}. The pipeline of generating one sample is independent with others. With such properties, it is easy to take advantage of parallel computing which makes the $V$-node expansion more efficient. Note that since $P(z|b, a)$ is no longer computed explicitly, the sampling frequency of the measurement in a $V$-node is applied instead to update its ancestors as shown in Algorithm~\ref{alg: qnode update} and~\ref{alg: vnode update}. Note that, besides forward sampling, the procedure of Bayes belief update can also be run in parallel with the posterior probablity of each state computed independently. The functions run on the GPU are shown as bold font in Algorithm~\ref{alg: qnode construction}.
\begin{algorithm}
  \caption{$V$-node Construction}
  \label{alg: vnode construction}
  \SetKwInOut{Input}{Input}
  \SetKwInOut{Output}{Output}

  \Input{$b$, $z$, $w$, $q$}
  \Output{$v$}

  $v.belief = b$\;
  $v.observation = z$\;
  $v.weight = w$\;
  $v.parent = q$\;
  $v.U_V = V_{FIB}(b)$\;
  $v.L_V = V_{PBVI}(b)$\;
  $v.H_V = v.U_V - v.L_V$\;
  $v.E_V = v$\;
\end{algorithm}
\begin{algorithm}
  \caption{$Q$-node Update}
  \label{alg: qnode update}

  $r = R(q.belief, q.action)$\;
  $q.U_Q = r + \sum_{v\in q.children} v.U_V \times v.weight$\;
  $q.L_Q = r + \sum_{v\in q.children} v.L_V \times v.weight$\;
  $v = \argmax_{v'\in q.children} \gamma \times v'.weight \times v'.heuristic$\;
  $q.H_Q = \gamma \times v.weight \times v.heuristic$\;
  $q.E_Q = v.E_V$\;
\end{algorithm}
\begin{algorithm}
  \caption{$V$-node Update}
  \label{alg: vnode update}

  $v.U_V = \max_{q\in v.children} q.U_Q$\;
  $v.L_V = \max_{q\in v.children} q.L_Q$\;
  $q = \argmax_{q'\in v.children} q'.H_Q \times \mathbf{1}_{q'.U_Q>v.U_V}$\;
  $v.H_V = q.H_Q$\;
  $v.E_V = q.E_Q$\;
\end{algorithm}

\section{Experiments}
\label{sec: experiments}
In this section, we will first specify the stochastic motion planning model, including each element in the tuple introduced in Sec.~\ref{sec: problem formulation}. Next, we show an simple example on a $5\times 5$ 2-D occupancy grid map to illustrate the behavior of QVTS. Finally, the proposed method is compared with the state-of-art motion planning methods which cannot handle state uncertainty without further simplication.

\begin{figure}[h]
  \centering
  \includegraphics[scale=0.11]{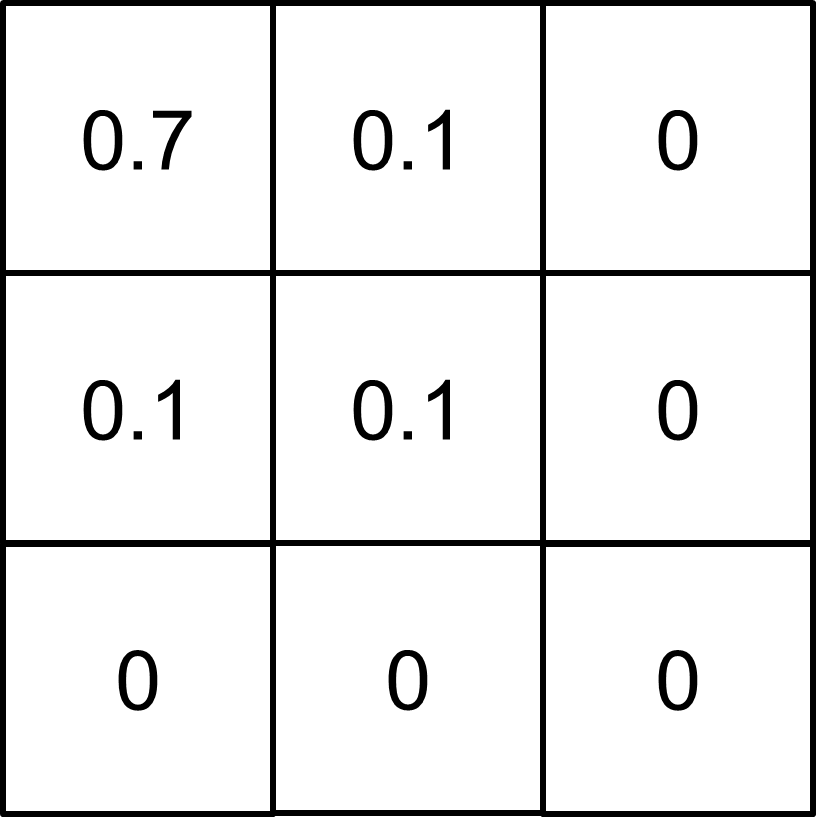}
  \includegraphics[scale=0.11]{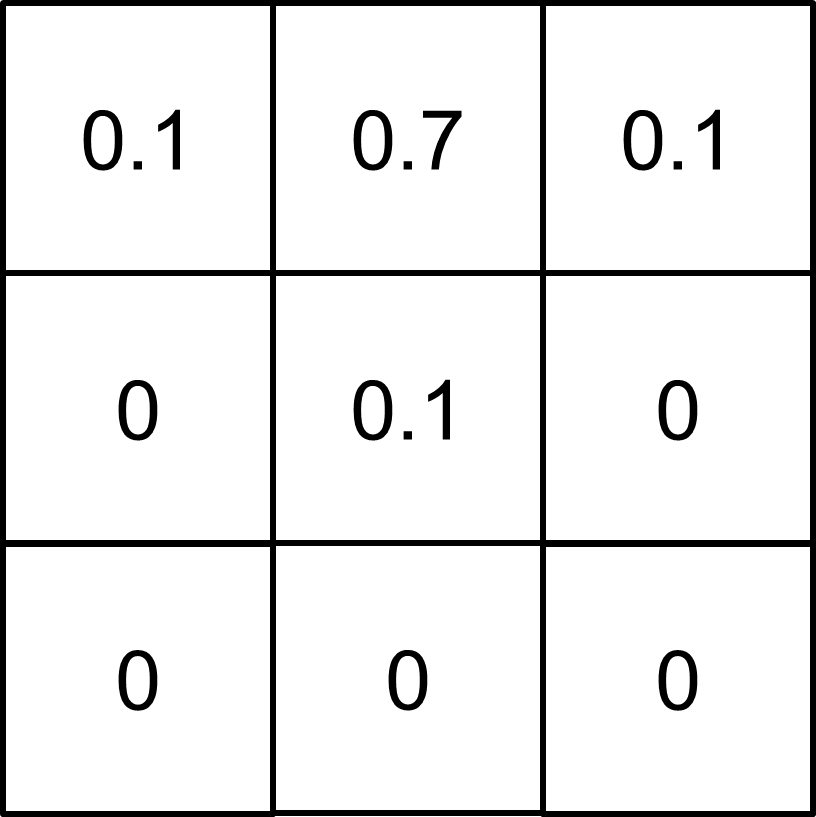}
  \includegraphics[scale=0.11]{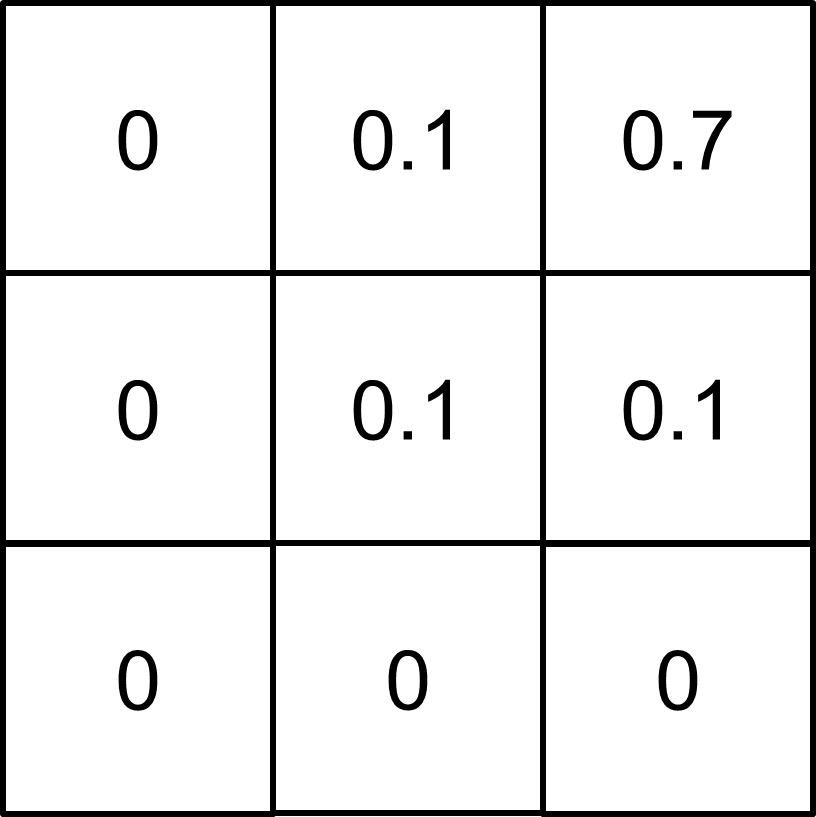}
  \includegraphics[scale=0.11]{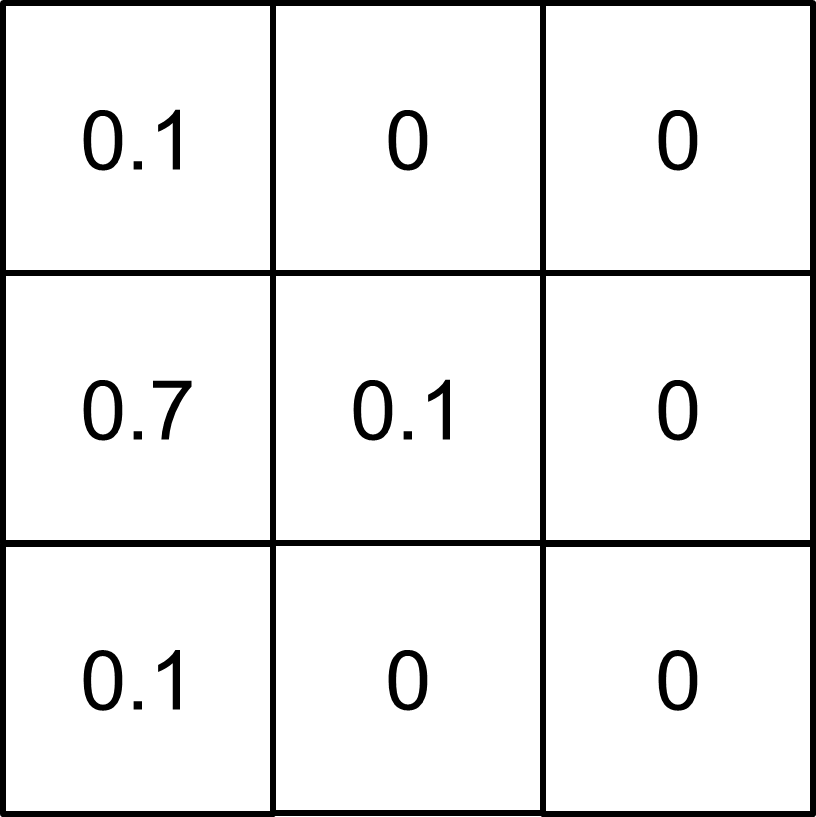}
  \includegraphics[scale=0.11]{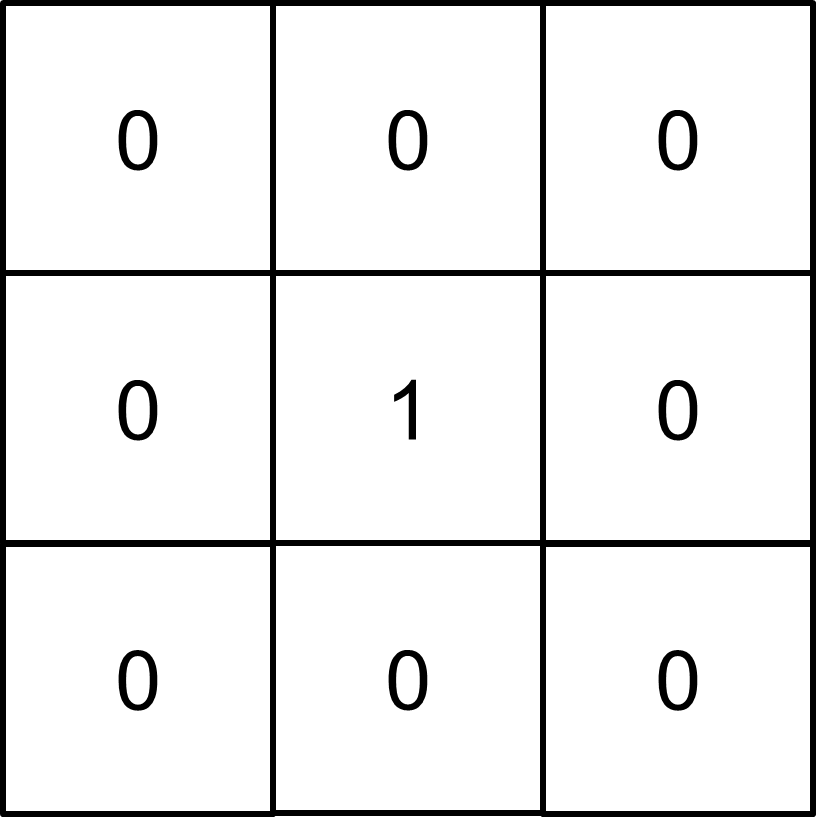} \\
  \smallskip
  \includegraphics[scale=0.11]{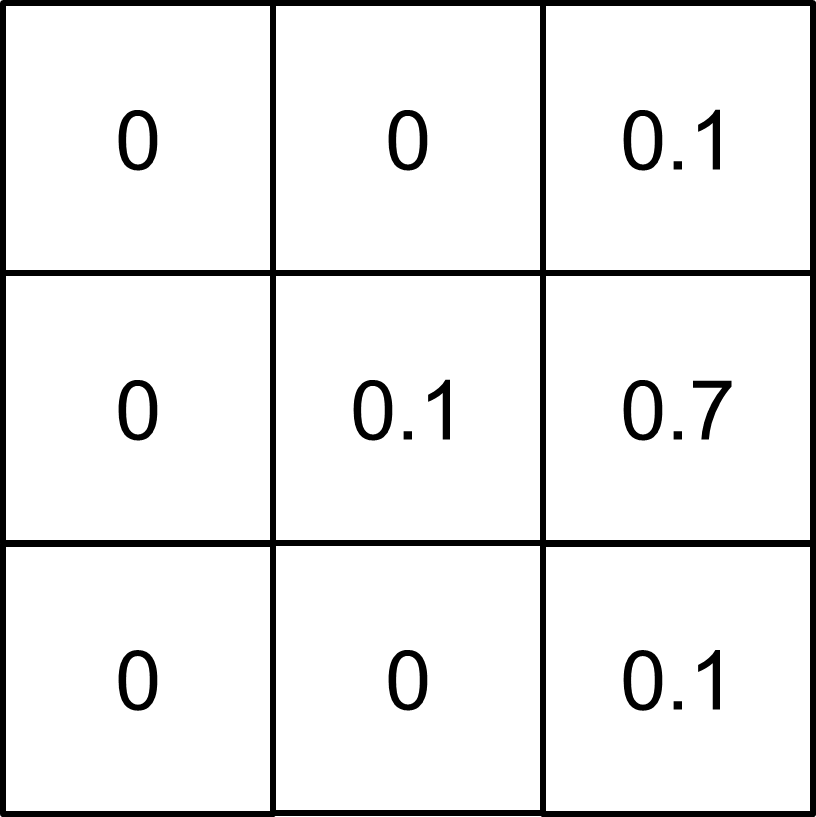}
  \includegraphics[scale=0.11]{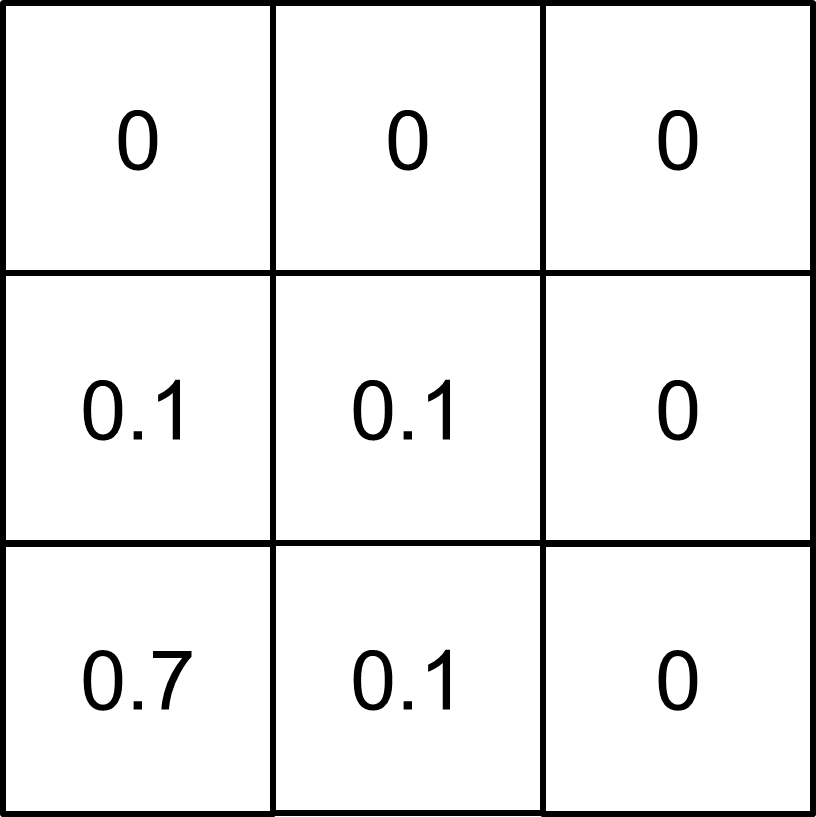}
  \includegraphics[scale=0.11]{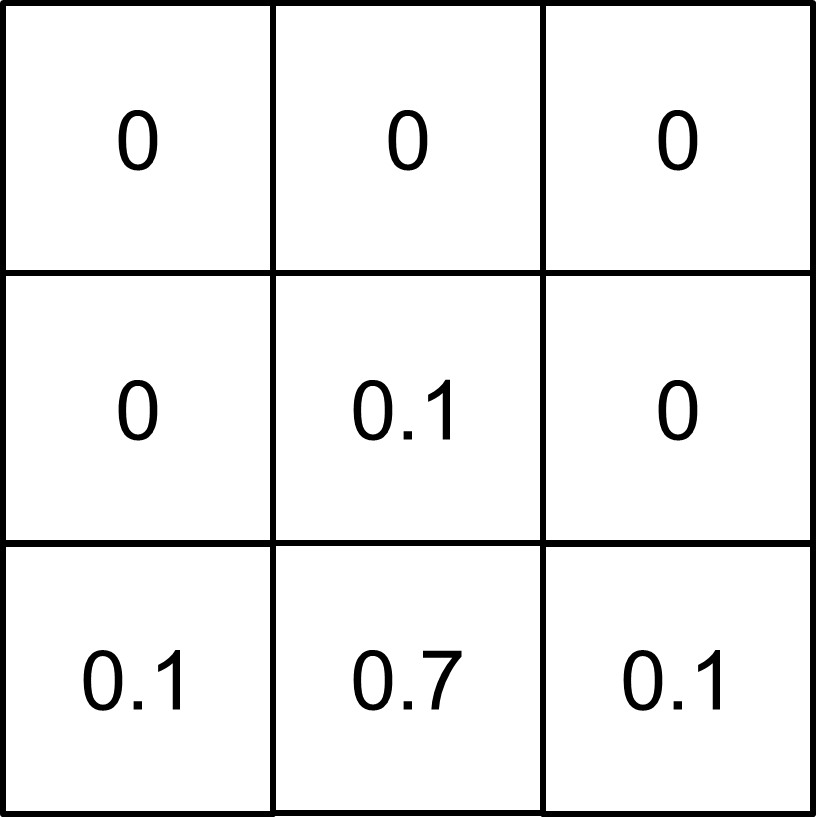}
  \includegraphics[scale=0.11]{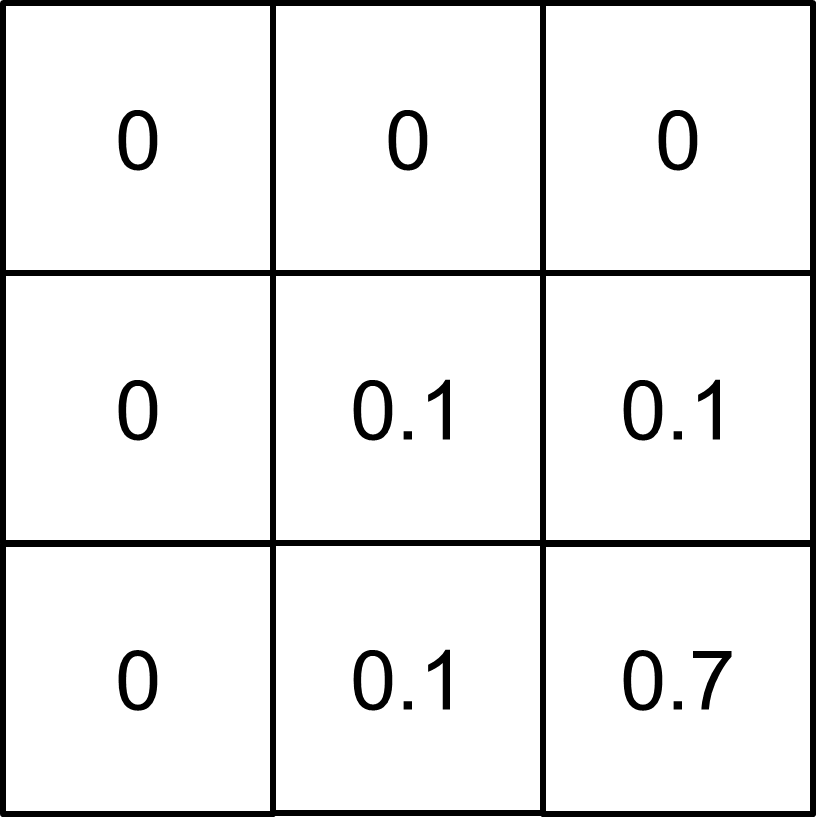}
  \includegraphics[scale=0.11]{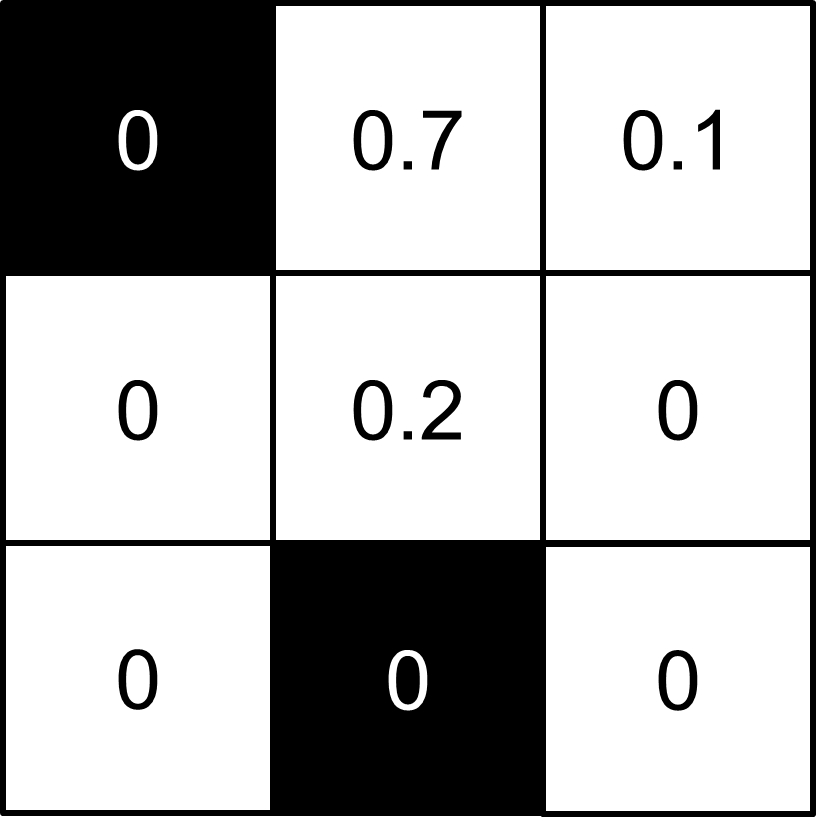}
  \caption{Assuming the robot is at the center, the first nine figures shows the transition probability, $T'(x, a, y)$, for each action assuming all cells in $\mathcal{N}(x)$ are free. The last figure shows an example of $T(x, a, y)$ for the upward action where some cells in $\mathcal{N}(x)$ are occupied. In this case, the probability assigned to the occupied cells in $T'(x, a, y)$ is accumulated to the current location of the robot.}
  \label{fig: model transition probability}
\end{figure}

On a deterministic occupancy grid map, a robot starting at a free cell $x$ is allowed to execute nine different actions $a\in\mathcal{A} = \{0, 1, \dots, 8\}$, moving to one of its nine neighbors, $\mathcal{N}(x)$, including the current location of the robot. To ease the description, each cell $N_a(x)\in\mathcal{N}(x)$ is indexed with the corresponding action, where $N_4(x)=x$. If all cells of $\mathcal{N}(x)$ are free, the transition probability $T'(x, a, y)$ is shown in Figure~\ref{fig: model transition probability}.
However, in the case that a cell $y$ is occupied, the probability assigned to $y$ is accumulated to $x$, which defines the true transition probability,
\begin{equation}
  \label{eq: transition probability with occupied cells}
  \begin{aligned}
    \text{For } m(y) &= 0:\\
    &T(x, a, y) = T'(x, a, y).\\
    \text{For } m(y) &= 1:\\
    &T(x, a, y) = 0, \\
    &T(x, a, x) = T'(x, a, x) + T'(x, a, y).
  \end{aligned}
\end{equation}

The robot is also assumed to have four sensors measuring the occupancy status of the neighbor cells, $\{N_1(x), N_3(x), N_5(x), N_7(x)\}$, i.e. $|\mathcal{Z}|=16$. Each sensor has the measurement likelihood $P(z_s = m(N_s(x)) | x) = 0.95$, where $s\in\{1, 3, 5, 7\}$.

The definition of the stage reward follows the principle that the reward assignment should drive the robot to the goal location. Define a function $r:\mathcal{X}\to\{-2, -1, 0\}$ such that,
\begin{equation*}
  r(x) =
  \begin{cases}
    -2, & \text{if } m(x) = 1. \\
    -1, & \text{if } m(x) = 0 \text{ and } x \text{ is not goal}. \\
    0, & \text{if } x \text{ is at goal}.
  \end{cases}
\end{equation*}
The goal location is assumed to be at a free cell implicitly. Since the actions are nondeterministic, the stage reward for an action also depends on the transition model,
\begin{equation*}
  R(x, a) =
  \begin{cases}
    -2, & \text{if } a = 4 \text{ and $x$ is not goal}. \\
    \sum_{y\in\mathcal{X}}r(y) T'(x, a, y), & \text{otherwise}.
  \end{cases}
\end{equation*}
The first case is to motivate the robot to move around instead of stopping at wrong locations. Note $T'$ is used instead of $T$ to distinguish actions moving into obstacles or free cells.

\subsection{Illustrative Example}
\label{subsec: qualitative experiment}
\begin{table*}[t]
  \centering
  \caption{Comparison between A*, MDP, and QVTS (averaged over $60$ independent experiments)}
  \label{tab: comparision of A*, mdp, and qvts}
  \renewcommand{\arraystretch}{1.2}
  \begin{tabular}{c | c | c | c | c | c}
    \hline
    & reward & collisions & steps & failure rate & planning time \\
    \hline
    A* & $-20.33\pm 0.37$ & $6.36\pm 3.39$ & $167.04\pm 29.08$ & $0.13$ & $4.61\pm 3.45$ms \\
    \hline
    MDP & $-20.06\pm 0.15$ & $1.58\pm 1.21$ & $175.82\pm 41.06$ & $0.07$ & $-$ \\
    \hline
    QVTS & $-20.07\pm 0.17$ & $1.36\pm 1.15$ & $132\pm 19.30$ & $0.02$ & $1376.49\pm 122.25$ms \\
    \hline
  \end{tabular}
\end{table*}
\begin{figure}[t]
  \centering
  \includegraphics[scale=0.12]{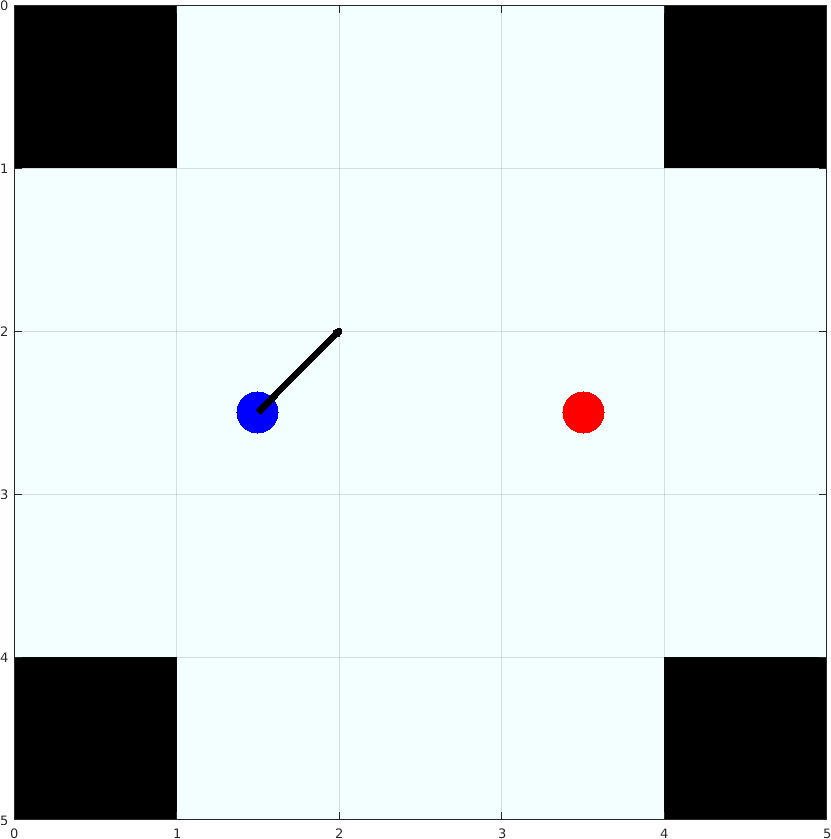}
  \includegraphics[scale=0.12]{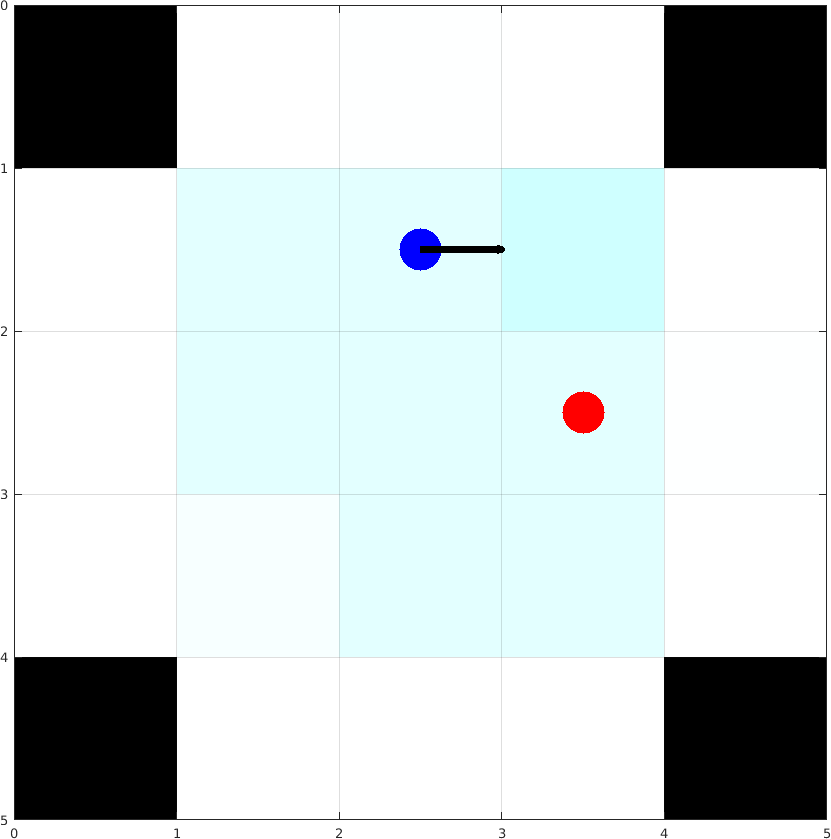}
  \includegraphics[scale=0.12]{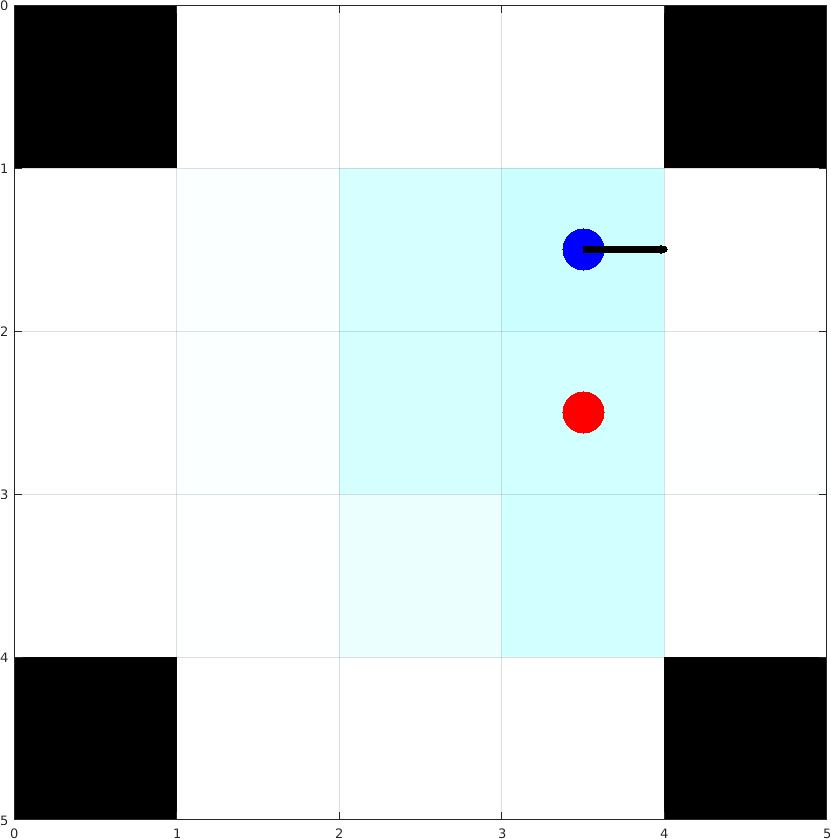}\\
  \smallskip
  \includegraphics[scale=0.12]{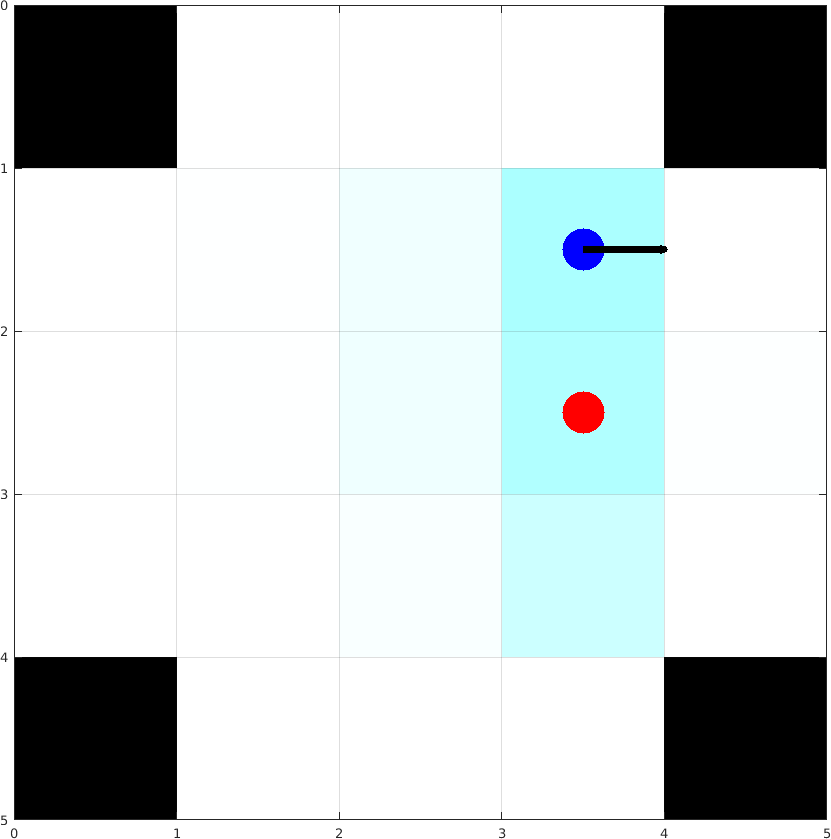}
  \includegraphics[scale=0.12]{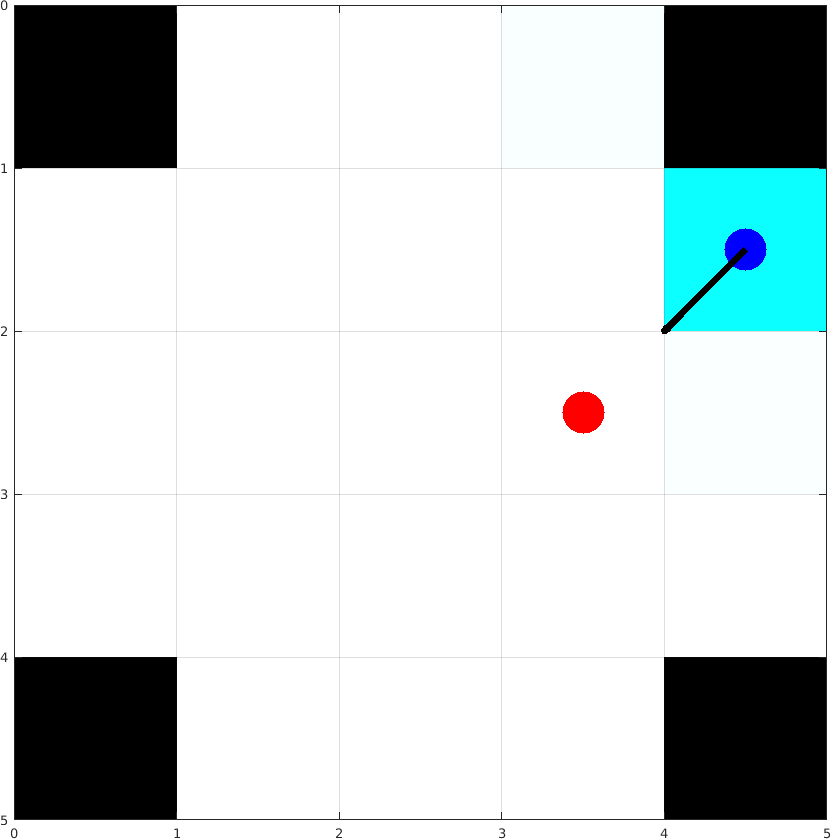}
  \includegraphics[scale=0.12]{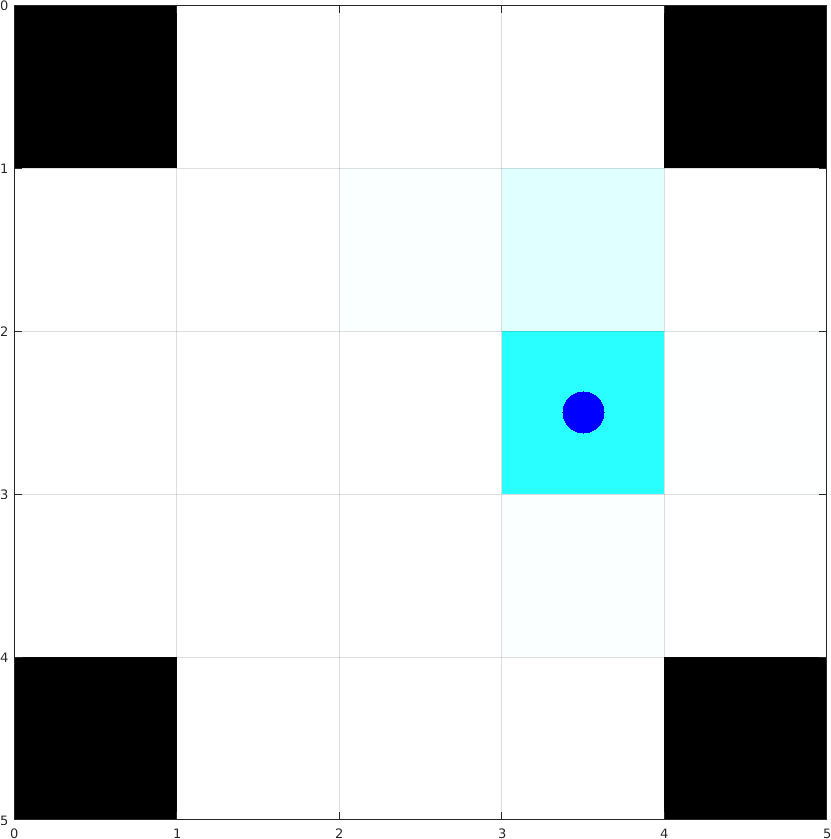}
  \caption{A robot (\textcolor{blue}{blue}) tries to move to the goal location (\textcolor{red}{red}) with uniform initial belief over the free cells. The arrow (\textcolor{black}{black}) represents the actions taken at each step. The belief of the robot is color (\textcolor{Turquoise}{cyan}) coded with darker cells representing higher propability. Note that, at the thrid step, although a right action is planned, the robot stops at the current position because of the motion stochasticity. However, the belief states changes accordingly.}
  \label{fig: path on a 5x5 map}
\end{figure}
Figure~\ref{fig: path on a 5x5 map} shows a simple example to illustrate the proposed QV-Tree Search. On a $5\times 5$ occupancy grid map, the task of the robot is to move to the goal location with uniform initial belief over the free cells. Since the robot can only localize itself around the four corners of the map, moving directly from start to the goal will result in high uncertainty of the robot's location. The QV-Tree Search chooses to move to the top right corner first and then go to the goal to increase the probability of completing the task. Although the final belief is not fully concentrated at the goal location, the QV-Tree Search decides to stop at the current position because of two reasons. First, further movement will not increase the probability of reaching the goal. The second reason is the ratio of the reward of a feasible motion and stopping at a wrong location. In the case that the cost of a wrong stop is much higher than that of a feasible motion, the robot will keep moving to increase the discounted reward summation over the infinite horizon.

\subsection{Quantitative Comparison}
\label{subsec: quantitative comparison}
\begin{figure}[htbp]
  \centering
  \includegraphics[scale=0.2]{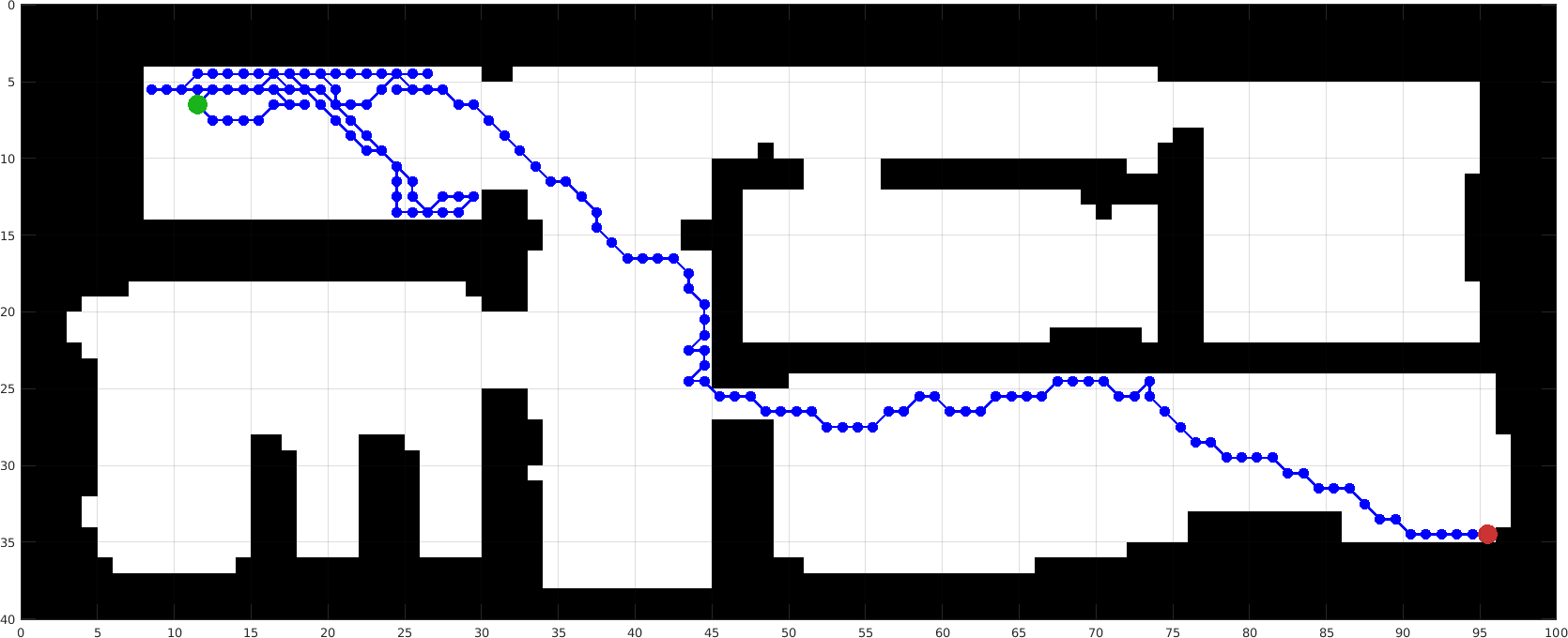}\\
  \includegraphics[scale=0.2]{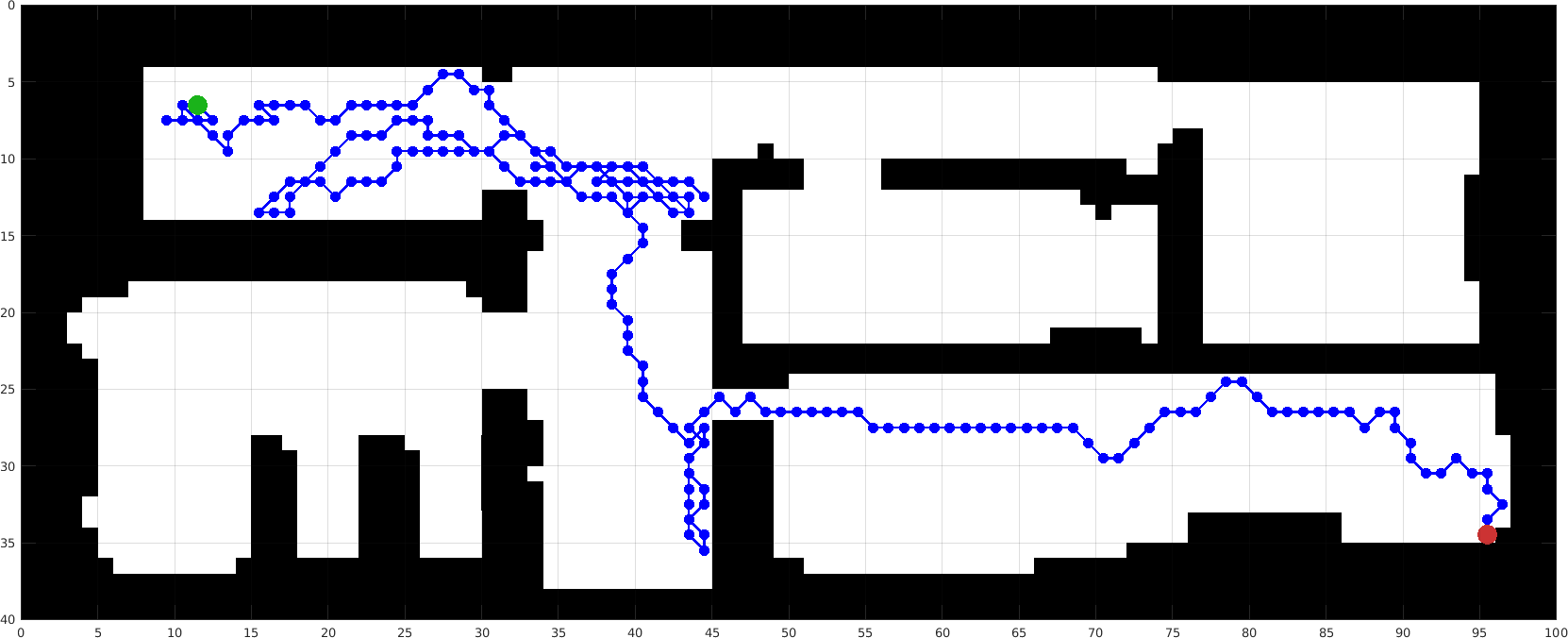}\\
  \includegraphics[scale=0.2]{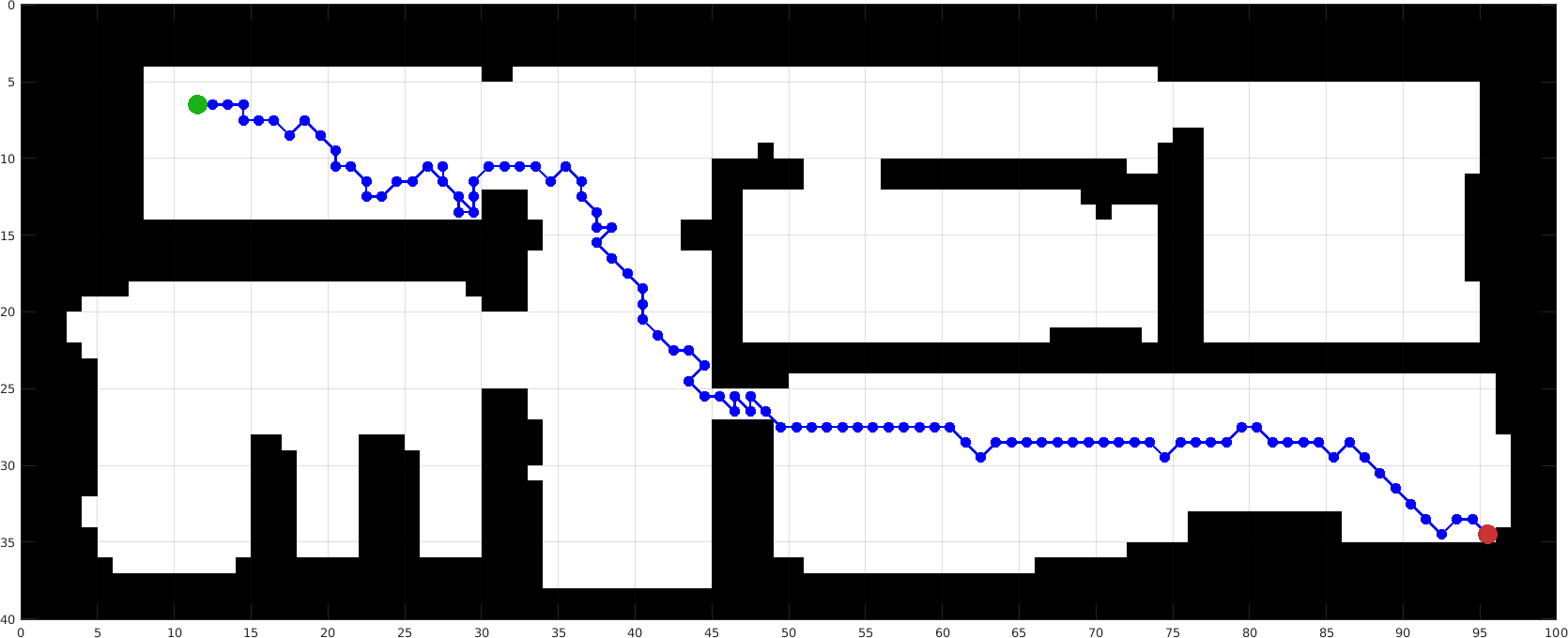}\\
  \caption{From top to bottom are the paths generated by A*, MDP, and QVTS in one of the experiments. The start and the goal are shown as \textcolor{OliveGreen}{green} and \textcolor{Maroon}{red} dots respectively.}
  \label{fig: path on a 100x40 map}
\end{figure}

To further evaluate the performance of the proposed algorithm, QVTS is compared with the state-of-art motion planning methods, A* and MDP, on a occupancy map of size $100\times 40$. As in Sec.\ref{subsec: qualitative experiment}, a robot is still required to reach the goal location with uniform initial belief over the free cells. The comparison includes summation of discounted reward, number of collisions with the environment, total number of steps reaching the goal, failure rate, and the average online planning time per step. The planning time is collected with Intel CPU Core i7-6600U and NVIDIA GPU M500M. Note that the planning time for MDP is omitted since the action for each step only requires one table lookup, whose time is neglectable. For A*, we use the open sourced implementation from~\cite{liu2017planning}. MDP is solved with self-implemented value iteration\footnote{\url{https://github.com/KumarRobotics/path_planning_2d}}. Since both A* and MDP require the location of the robot to be known, the mode of the belief is used in the experiments for these two methods. The statistics of the three methods is shown in Table~\ref{tab: comparision of A*, mdp, and qvts} averaged over $60$ independent experiments for each method. Figure~\ref{fig: path on a 100x40 map} shows example paths generated with the three methods.

According to Table~\ref{tab: comparision of A*, mdp, and qvts}, all methods share similar reward summation due to the fact that the reward of later steps is heavliy discounted. A* has slightly lower reward because it generates more collisions in the early stages comparing with the other two methods. The fewer collisions of the other two methods is because they are able to actively avoid the obstacles considering the stochastic dynamics of the robot. The proposed QVTS is significantly advantageous in success rate and total number of steps. Such advantages of QVTS come from the capability of active localization. Figure~\ref{fig: path on a 100x40 map} shows that the proposed method can localize the robot within the first $20-30$ steps, which helps the robot reach the goal with higher probability and fewer steps. More details on the path and belief evolution of the three methods can be found in the supplementary video.

\section{Conclusion}
\label{sec: conclusion}
In this paper, we consider the problem of discrete motion planning for a robot with finite state, action, and observation space. In order to handle the dynamics and measurement noise, QVTS is proposed combining both offline and online POMDP approximation methods. In order to the reduce the online processing time, only a limited number of possible observations are considered during a node expansion. Also, by ignoring most of the unlikely observations, the proposed QVTS is able to handle a large scale observation set. Experiments show that the proposed method is more robust and efficient comparing with the traditional methods.

However, the proposed QVTS is still limited in the following two aspects. First, QVTS is not able to handle continous system directly, which is usually natural in modeling robot dynamics and sensor measurements. Also, the proposed method cannot handle the large state spaces in exploration problems where the system state should include both the robot state and the environment. The underlying reason for both of the limitations is that the belief of the state can no longer be represented as a finite-size vector. It follows that the Bayes belief update, as a necessary step in QVTS, is no longer feasible. In the future work, we will try to address this problem by reprenting the belief with state samples.

\addtolength{\textheight}{-4.5cm}



\bibliographystyle{IEEEtran}
\bibliography{ref}

\end{document}